\pdfoutput=1

\documentclass[11pt,a4paper]{article}
\usepackage[hyperref]{naacl2021}
\usepackage{times}
\usepackage{latexsym}
\usepackage{soul}
\usepackage{url}
\usepackage[utf8]{inputenc}
\usepackage[font=small]{caption}
\usepackage{graphicx}
\usepackage{amsmath}
\usepackage{amsthm}
\usepackage{booktabs}
\usepackage{algorithm}
\usepackage{algorithmic}
\usepackage{multirow}

\usepackage{ulem}
\usepackage{paralist}

\title{Modeling Diagnostic Label Correlation for Automatic ICD Coding}

\author{Shang-Chi Tsai\thanks{\hspace*{0.5em}Equal contribution.} \quad Chao-Wei Huang\footnotemark[1]\quad Yun-Nung Chen\\
National Taiwan University, Taipei, Taiwan\\
\texttt{$\{$d08922014,f07922069$\}$@csie.ntu.edu.tw \quad y.v.chen@ieee.org}}

\begin{document}
\maketitle
\begin{abstract}
Given the clinical notes written in electronic health records (EHRs), it is challenging to predict the diagnostic codes which is formulated as a multi-label classification task.
The large set of labels, the hierarchical dependency, and the imbalanced data make this prediction task extremely hard.
Most existing work built a binary prediction for each label independently, ignoring the dependencies between labels.
To address this problem, we propose a two-stage framework to improve automatic ICD coding by capturing the label correlation.
Specifically, we train a label set distribution estimator to rescore the probability of each label set candidate generated by a base predictor. 
This paper is the first attempt at learning the label set distribution as a reranking module for medical code prediction.
In the experiments, our proposed framework is able to improve upon best-performing predictors on the benchmark MIMIC datasets.
\footnote{The source code of this project is available at https://github.com/MiuLab/ICD-Correlation.}
\end{abstract}

\section{Introduction}
Clinical notes from electronic health records (EHRs) are free-from text generated by clinicians during patient visits.
The associated diagnostic codes from the International Classification of Diseases (ICD) represent diagnostic and procedural information of the visit.
The ICD codes provide an standardized and systematic way to encode information and has several potential use cases~\cite{choi2016doctor}.

\begin{figure}[t!]
\centering
\includegraphics[width=\linewidth]{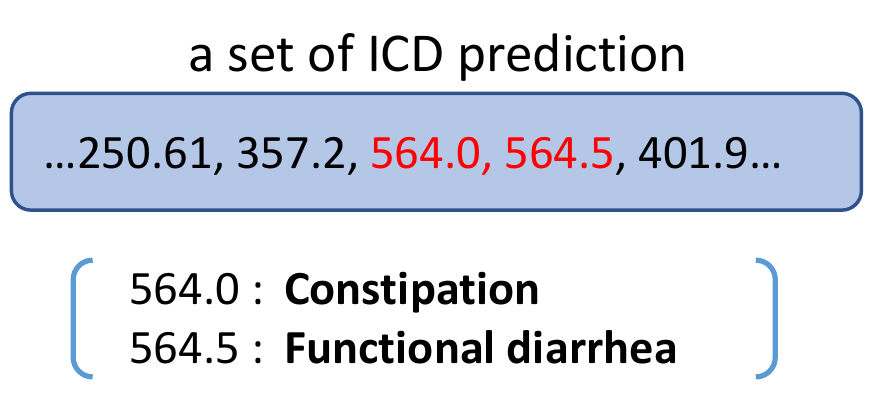}
\caption{An example of conflicting predictions. While these two codes share the same root in the hierarchical ICD structure and are semantically similar, they are unlikely to appear together.}
\label{fig:error-example}
\end{figure}

Considering that manual ICD coding has been shown to be labor-intensive~\cite{o2005measuring}, several approaches for automatic ICD coding has been proposed and investigated by the research community~\cite{perotte2013diagnosis,kavuluru2015empirical}.
With recent introduction of deep neural networks, the performance of automatic ICD coding has been improved significantly~\cite{choi2016doctor,shi2017towards,mullenbach-etal-2018-explainable,baumel2018multi,xie-xing-2018-neural,li2020multirescnn,ijcai2020-461,cao-etal-2020-hypercore}.
Prior work on neural models mostly treated the task of automatic ICD coding as a multi-label classification problem.
These models mostly employ a shared text encoder, and build one binary classifier for each label on top of the encoder.
This architecture along side with binary cross-entropy loss make the prediction of each label independent of each other, which might lead to incomplete or conflicting predictions.
An example of such error is shown in Figure~\ref{fig:error-example}.
This issue is especially problematic in ICD code prediction, since the ICD codes share a hierarchical structure. That is, the low-level codes are more specific, and the high-level ones are more general.
In some cases, the low-level codes under the same high-level category are more likely to be jointly diagnosed. Rare codes also have more opportunity to be considered from the frequent codes in the same high-level class.
Prior work considered the hierarchical dependencies between ICD codes by using hierarchical SVM~\cite{perotte2013diagnosis} or by introducing new loss terms to leverage the ICD structure~\cite{tsai-etal-2019-leveraging}.
However, they borrowed the dependency from domain experts and did not consider the label correlation in the data.

Inspired by the success of reranking techniques on automatic speech recognition~\cite{ostendorf-etal-1991-integration} and dependency parsing~\cite{zhu-etal-2015-ranking,sangati-etal-2009-generative}, we propose a two-stage reranking framework for ICD code prediction, which captures the label correlation without any expert knowledge.
In the first stage, we use a base predictor to generate possible label set candidates.
In the second stage, a label set reranker is employed to rerank the candidates.
We design two rerankers to help to capture the correlation between labels.
The experimental results show that our proposed framework consistently improves the results of different base predictors on the benchmark MIMIC datasets~\cite{saeed2011multiparameter,johnson2016mimic}.
The results also show that the proposed framework is model agnostic, i.e., we can use any base predictor in the first stage.

Data privacy is a major difficulty for medical NLP research.
The personal health information (PHI) which explains a patient's ailments, treatments and outcomes is highly sensitive, making it hard to distribute due to privacy concerns.
In addition, EHRs across multiple hospitals or languages may contain different writing style, typos and abbreviations.
It is labor-demanding to train separate models for each hospital with their in-house data only.
The advantage of our proposed two-stage framework is that we can train base predictors with in-house data, while enjoying the universality of ICD codes to train a reranker on ICD labels from different sources. 
This reranker is able to generally work with various base predictor trained on health records from any specific hospital.

The contributions of this paper are 3-fold:
\begin{compactitem}
    \item This paper is the first attempt to improve multi-label classification with a reranking method for automatic ICD coding.
    \item The experiments show that the proposed approaches are capable of improving best-performing base predictors on the benchmark datasets MIMIC-2 and MIMIC-3, demonstrating its great generalizability.
    \item The proposed framework has the great potential of benefiting from extra ICD labels, reducing the demand of paired training data towards scalability in the medical NLP field.
\end{compactitem}

\section{Related Work}
This paper focuses on multi-label medical code prediction; hence, We briefly describe the related background about medical code prediction and multi-label classification.

\subsection{Medical Code Prediction}
ICD code prediction is a challenging task in the medical domain. It has been studied since 1998 \cite{de1998hierarchical} and several recent work attempted to approach this task with neural models.
\citet{choi2016doctor} and \citet{baumel2018multi} used recurrent neural networks (RNN) to encode the EHR data for predicting diagnostic results.
\citet{li2020multirescnn} recently utilized a multi-filter convolutional layer and a residual layer to improve the performance of ICD prediction.
On the other hand, several work tried to integrate external medical knowledge into this task.
In order to leverage the information of definition of each ICD code, RNN and CNN were adopted to encode the diagnostic descriptions of ICD codes for better prediction via attention mechanism~\cite{shi2017towards,mullenbach-etal-2018-explainable}.
Moreover, the prior work tried to consider the hierarchical structure of ICD codes~\cite{xie-xing-2018-neural}, which proposed a tree-of-sequences LSTM to simultaneously capture the hierarchical relationship among codes and the semantics of each code.
Also, \citet{tsai-etal-2019-leveraging} introduced various ways of leveraging the hierarchical knowledge of ICD by adding refined loss functions.
Recently, \citet{cao-etal-2020-hypercore} proposed to train ICD code embeddings in hyperbolic space to model the hierarchical structure. Additionally, they used graph neural network to capture the code co-occurrences.

\begin{figure*}[t!]
\centering
\includegraphics[width=\linewidth]{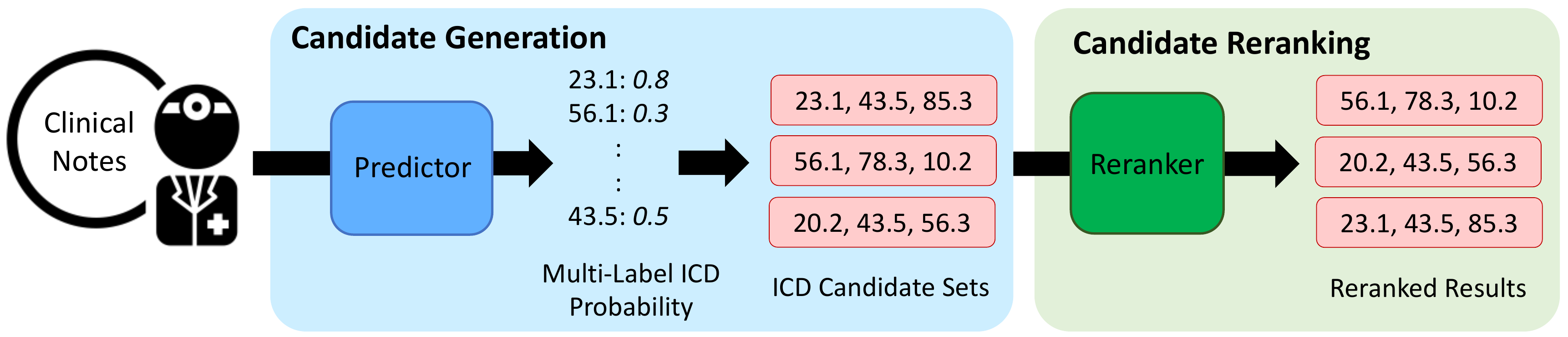}
\caption{Illustration of our proposed framework.}
\vspace{-3mm}
\label{fig:framework}
\end{figure*}

\subsection{Multi-Label Classification}
Multi-label classification problems are of broad interest to the machine learning community.
The goal is to predict a subset of labels associated with a given object.
One simple solution to a multi-label classification problem is to transform the problem into $n$ binary classification problems, where $n$ denotes the number of labels.

This approach makes an assumption that the predictions of each label are independent.
However, in practice, the labels are usually dependent, making these predictors produce undesired predictions.

There are numerous methods developed to alleviate this issue.
\citet{read2009classifier} proposed classifier chains (CC), which introduce sequential dependency between predictions by adding the decision of one classifier to the input of the next classifier.
\citet{cheng2010bayes} generalized CC to probabilistic classifier chains (PCC), where the proposed approach estimates the joint probability of labels and provides a proper interpretation of CC.
Nevertheless, the recurrence between classifiers makes these methods less efficient and not applicable to tasks with large amount of labels.

Another line of research has leveraged the label dependencies that are known beforehand.
\citet{deng2014large} used label relation graphs for object classification.
\citet{tsai-etal-2019-leveraging} utilized the hierarchical structure of ICD codes to improve the ICD code prediction.
These methods relied on known structures of the labels, which may not be easily accessible and less general.

Some prior work tried to learn label correlation and dependencies directly from the dataset.
\citet{zhang2018multi} introduced residual blocks to capture label correlation. This method requires paired training data, while our framework can learn from ICD codes only.

The concept of \emph{retrieve-and-rerank} has been widely used in automatic speech recognition~\cite{ostendorf-etal-1991-integration}, natural language processing~\cite{collins2005discriminative} and machine translation~\cite{shen2004discriminative}.
\citet{li2019learning} proposed to rerank the possible predictions generated by a base predictor with a calibrator.
This method is conceptually similar to our framework, where we both follow the retrieve-and-rerank procedure.
The main difference between is that they leveraged an extra dataset for training the calibrator, while we train a distribution estimator on the same dataset as our base predictor.

\begin{figure*}[t!]
\centering
\includegraphics[width=\linewidth]{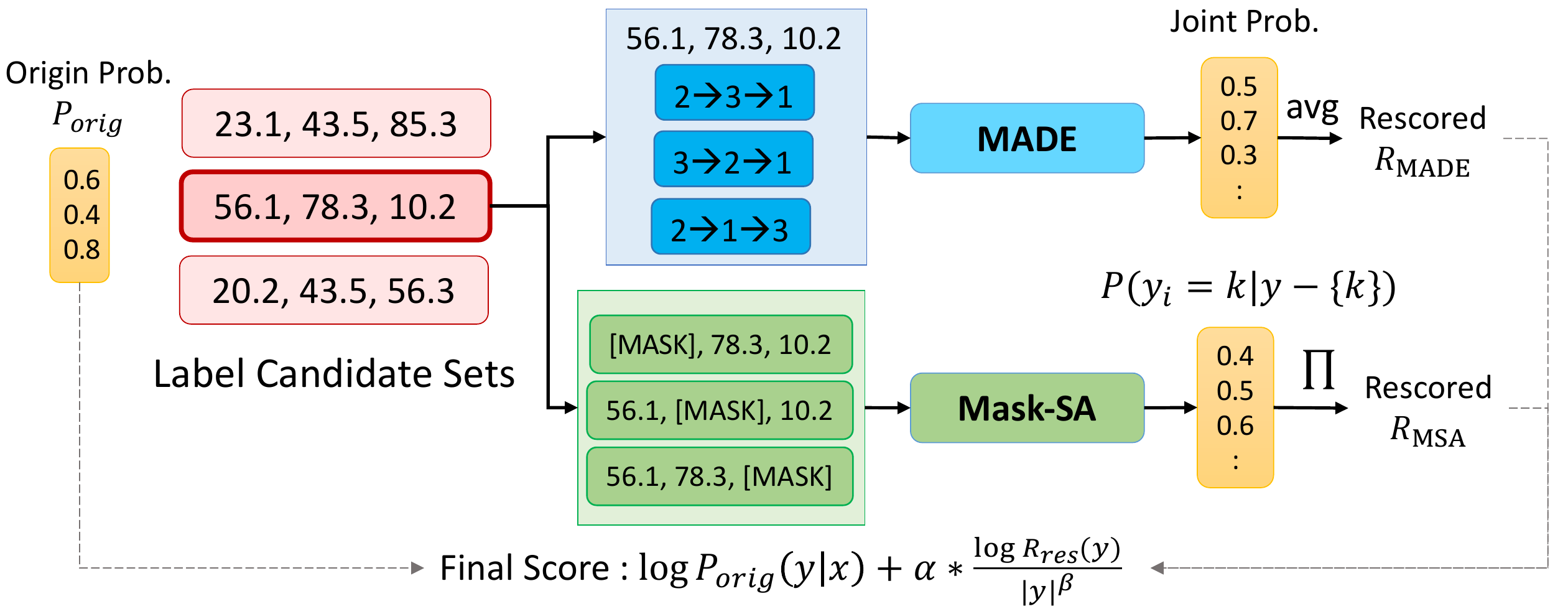}
\caption{Illustration of the proposed reranking process.}
\label{fig:rescore}
\end{figure*}

\section{Proposed Framework}

The task of ICD code prediction is usually framed as a multi-label classification problem~\cite{kavuluru2015empirical,mullenbach-etal-2018-explainable}.
Given a clinical record $\mathbf{x}$ in EHR, the goal is to predict a set of ICD codes $\mathbf{y} \subseteq \mathcal{Y}$, where $\mathcal{Y}$ denotes the set of all possible codes.
This subset is typically represented as a binary vector $\mathbf{y} \in \{0,1\}^{|\mathcal{Y}|}$, where each bit $y_i$ indicates the presence or absence of the corresponding label.

The proposed framework is illustrated in Figure~\ref{fig:framework} and consists of two stages: 
\begin{compactenum}
    \item \textbf{Label set candidate generation} provides multiple ICD set candidates through a base binary predictor, which is detailed in Section~\ref{sec:cand-gen}.
    \item \textbf{Label set candidate reranking} estimates the probability by leveraging the label correlation for reranking the candidates, which is detailed in Section~\ref{sec:cand-rerank}.
\end{compactenum}

\subsection{Candidate Generation}
\label{sec:cand-gen}
In the first stage of the framework, we employ a base predictor to perform probabilistic prediction for all labels, and we use the predicted probabilities to generate top-$k$ most probable label sets.
More formally, given a clinical note $\mathbf{x}$, we perform a base predictor and obtain the prediction for all labels:
\begin{equation*}
    P_\text{base}(y_i = 1 \mid \mathbf{x}, \theta_\text{base}), \quad i=1,2,\cdots,|\mathcal{Y}|,
\end{equation*}
where $\theta_\text{base}$ denotes the parameters of the base predictor.
The predicted results are used to generate top-$k$ probable sets, i.e., $\hat{\mathbf{y}} \subseteq \mathcal{Y}$ with top-$k$ highest probability prediction:
\begin{equation*}
    P_\text{base}(\hat{\mathbf{y}} \mid \mathbf{x}, \theta_\text{base}) = \prod_{i=1}^{|\mathcal{Y}|}{P_\text{base}(y_i = \hat{\mathbf{y}_i} \mid \mathbf{x}, \theta_\text{base})}.
\end{equation*}
Although there are $2^{|\mathcal{Y}|}$ possible subsets, the top-$k$ sets can be efficiently generated with dynamic programming as described in the prior work~\cite{li2016conditional}.

\subsection{Candidate Reranking}
\label{sec:cand-rerank}
One drawback of the base predictor is the assumption about independent labels.
To address this issue, in the second stage of the framework, we introduce a label set reranker to rerank the label set candidates generated in the previous stage.
The reranker is designed to capture correlation and co-occurrence between labels.
Given a label set candidate $\hat{\mathbf{y}}$, a reranker should be able to provide a reranking score $R(\hat{\mathbf{y}})$, where higher score indicates that the label set is more probable to appear.
Similar to prior work~\cite{zhu-etal-2015-ranking}, We rerank the candidates according to their new scores defined as
\begin{equation*}
    \log P_\text{base}(\hat{\mathbf{y}} \mid \mathbf{x}, \theta_{base}) + \alpha \cdot R(\hat{\mathbf{y}}),
\end{equation*}
where $\alpha$ is a hyperparameter.
We use the label set with the highest score after reranking as the final prediction.
Note that that reranking is done on label sets, not individual labels.

We employ two rerankers and describe them in the following subsections.
Note that we do not restrict the design of rerankers to those we proposed; one can design their own reranker and plug it into the proposed framework.
Our reranking framework is illustrated in Figure~\ref{fig:rescore}.

\subsubsection{MADE Reranker}
One intuitive way to assign scores to $\hat{\mathbf{y}}$ is using the joint probability $P(\hat{\mathbf{y}})$.
Higher joint probability indicates that the label set is more probable to appear, which aligns with our requirement to rerankers.
However, the joint probability $P(\hat{\mathbf{y}})$ is often intractable.
Therefore, we can only make an estimation with a density estimator.

Here we employ a masked autoencoder (MADE)~\cite{germain2015made} as the density estimator.
MADE estimates the joint probability of a binary vector $P(\hat{\mathbf{y}})$ by decomposing it in an autoregressive fashion with a random ordering
\begin{equation*}
    P_\text{MADE}(\hat{\mathbf{y}}) = \prod_{i=1}^{|\mathcal{Y}|}{P_\text{MADE}(y_i = \hat{\mathbf{y}}_i \mid \hat{\mathbf{y}}_{o<i}, \theta_\text{MADE})},
\end{equation*}
where $o$ denotes a random permutation of $\{1, 2, \cdots, |\mathcal{Y}|\}$, $o(i)$ denotes the new ordering of $i$, $\hat{\mathbf{y}}_{o<i} = \{\hat{\mathbf{y}}_j \mid o(j)<o(i)\}$ denotes the set of all elements precede $\hat{\mathbf{y}}_i$ in the new ordering, and $\theta_{MADE}$ denotes parameters of the MADE model.
MADE introduces a sequential dependency between labels.
It enforces this dependency by masking certain connections in the multi-layer perceptron, making the output of $y_i$ only depends on $\hat{\mathbf{y}}_{o<i}$.

The MADE model is trained with the labels from the training set of the ICD code prediction task.
We use stochastic gradient descent to optimize the parameters $\theta_\text{MADE}$, and the training objective is to minimize the binary cross-entropy loss $\mathcal{L}(\hat{\mathbf{y}})$:
\begin{align*}
    -&\frac{1}{|\hat{\mathbf{y}}|} \sum_{i=1}^{|\hat{\mathbf{y}}|}
     \Big( \hat{\mathbf{y}}_i \log P_\text{MADE}(y_i = 1 \mid \hat{\mathbf{y}}_{o<i}, \theta_\text{MADE})  \\
    +& (1-\hat{\mathbf{y}}_i) \log P_\text{MADE}(y_i = 0 \mid \hat{\mathbf{y}}_{o<i}, \theta_\text{MADE}) \Big).
\end{align*}

Because we do not know which ordering performs the best, we can sample $n$ different orderings and use the ensemble of these orderings to improve estimation:
\begin{align*}
    P_\text{MADE}(\hat{\mathbf{y}}) &=\\ \frac{1}{n}\sum_{j=i}^{n}&\prod_{i=1}^{|\mathcal{Y}|}{P_\text{MADE}(y_i = \hat{\mathbf{y}}_i \mid \hat{\mathbf{y}}_{o_j<i}, \theta_\text{MADE})}.
\end{align*}
The illustration can be found in the blue box of Figure~\ref{fig:rescore}.

Given a label set candidate $\hat{\mathbf{y}}$, we define the score $R_\text{MADE}(\hat{\mathbf{y}})$ as
\begin{equation*}
    R_\text{MADE}(\hat{\mathbf{y}}) = \frac{\log P_\text{MADE}(\hat{\mathbf{y}})}{|\hat{\mathbf{y}}|^{\beta}},
\end{equation*}
where $|\hat{\mathbf{y}}|$ denotes the size of the subset $\hat{\mathbf{y}}$, and $\beta$ is a hyperparameter.
$|\hat{\mathbf{y}}|^{\beta}$ serves as a length penalty similar to the one used in sequence generation~\cite{wu2016google}.
We find that this length penalty is crucial to the reranker, and without it the score would favor subsets with smaller size.


\subsubsection{Masked Self-Attention Reranker (Mask-SA)}
As described in the previous subsection, MADE uses a sequential factorization to estimate the joint probability of a label set.
This formulation forces the prediction of $y_i$ to only condition on a subset of inputs $\hat{\mathbf{y}}_{o<i}$.
With this restriction, the MADE model may fail to capture some crucial dependencies.

Inspired by the masked language modeling objective~\cite{devlin-etal-2019-bert}, we propose a masked self-attention reranker (Mask-SA).
Mask-SA takes as input a set of predicted labels $\hat{\mathbf{y}} \subseteq \mathcal{Y}$, which is the set representation of the predicted labels. It employs a cloze-style prediction method, where we mask one input at a time and ask the model to predict the masked input.
The advantage of this prediction method is that the output is conditioned on all inputs except for itself, which solves the restriction of the MADE model.
This procedure is very similar to a denoising autoencoder~\cite{vincent2008extracting}.
The illustration can be found in the green box of Figure~\ref{fig:rescore}.

The Transformer architecture~\cite{vaswani2017attention} has been shown to be efficient and effective in language modeling~\cite{dai-etal-2019-transformer}.
We use it as the architecture of the Mask-SA model, with a slight modification where we remove the positional encodings due to the fact that the predicted ICD codes have no sequential order.

More formally, Mask-SA estimates a distribution over the label vocabulary for the masked input given all other elements in the set $P_\text{MSA}(\hat{\mathbf{y}}_i \mid \hat{\mathbf{y}} - \{\hat{\mathbf{y}}_i\}, \theta_{MSA})$, where $\theta_{MSA}$ denotes the parameters of the Mask-SA model.
$\theta_\text{MSA}$ can be optimized with stochastic gradient descent to minimize the cross-entropy loss function.
Given a label set candidate $\hat{\mathbf{y}}$, we compute the score $R_\text{MSA}(\hat{\mathbf{y}})$ as
\begin{equation*}
    R_\text{MSA}(\hat{\mathbf{y}}) = \frac{\text{log } \prod_{i=1}^{|\hat{\mathbf{y}}|}{P_\text{MSA}(\hat{\mathbf{y}}_i \mid \hat{\mathbf{y}} - \{\hat{\mathbf{y}}_i\}, \theta_\text{MSA})}}{|\hat{\mathbf{y}}|^\beta},
\end{equation*}
where $\beta$ is a hyperparameter.
Note that in this formulation, the product of the conditional probabilities is not an exact estimation of the joint probability of $\hat{\mathbf{y}}$, but an analogy to the factorization made in the MADE model.

\section{Experiments}
In order to evaluate the effectiveness of our proposed framework, we conduct experiments on two benchmark datasets. We employ three different base predictors to validate the generalizability of the proposed framework.

\begin{table*}[t!]
\resizebox{\textwidth}{!}{ 
\centering
\begin{tabular}{lcccc|cccc}
\toprule
 \multirow{2}{*}{\bf Model} & \multicolumn{2}{c}{\bf Top-50 Dev} & \multicolumn{2}{c|}{\bf Top-50 Test} & \multicolumn{2}{c}{\bf All Dev} & \multicolumn{2}{c}{\bf All Test} \\
 \cmidrule{2-5}
 \cmidrule{6-9}
 &\bf MacroF & \bf MicroF & \bf MacroF & \bf MicroF &\bf MacroF & \bf MicroF & \bf MacroF & \bf MicroF \\
\midrule
CAML~\shortcite{mullenbach-etal-2018-explainable} & 54.03 & 61.76  & 53.46 & 61.41 & 7.70 & 54.29 & 8.84 & 53.87 \\
~+ MADE  & \bf 56.91$^\dag$ & \bf 62.31$^\dag$  & \bf 56.64$^\dag$ & \bf 62.40$^\dag$ & 7.79$^\dag$ & \bf 54.70$^\dag$ & 9.11$^\dag$ & \bf 54.29$^\dag$ \\
~+ Mask-SA & 56.57$^\dag$ & 62.14$^\dag$  & 56.33$^\dag$ & 62.26$^\dag$ & \bf 8.06$^\dag$ & 54.53$^\dag$ & \bf 9.27$^\dag$ & 54.09$^\dag$\\
\midrule
MultiResCNN~\shortcite{li2020multirescnn} & 60.77 & 66.98  & 60.84 & 66.78 & 7.38 & 56.05  & 8.50 & 55.31\\
~+ MADE & \bf 62.10$^\dag$ & 67.13$^\dag$  & 62.00$^\dag$ & 67.13$^\dag$ & 7.75$^\dag$ & 57.08$^\dag$  & 8.81$^\dag$ & 56.21$^\dag$ \\
~+ Mask-SA & 61.52$^\dag$ & \bf 67.15$^\dag$  & \bf 62.06$^\dag$ & \bf 67.22$^\dag$ & \bf 7.97$^\dag$ & \bf 57.12$^\dag$  & \bf 9.28$^\dag$ & \bf 56.49$^\dag$ \\
\midrule
LAAT~\shortcite{ijcai2020-461} & 65.53 & \bf70.38 & 65.05 & 70.01 & 7.48 & 57.18  & 8.74 & 56.56 \\ 
~+ MADE  & 65.92$^\dag$ & 70.34 & 65.29$^\dag$ & 70.13$^\dag$ & 7.92$^\dag$ & \bf 57.80$^\dag$ & 9.16$^\dag$ & \bf 57.26$^\dag$ \\ 
~+ Mask-SA & \bf 66.10$^\dag$ & 70.30 & \bf 65.44$^\dag$ & \bf 70.15$^\dag$ & \bf 8.08$^\dag$ & 57.76$^\dag$ & \bf 9.41$^\dag$ & 57.23$^\dag$ \\ 
\bottomrule
\end{tabular}
}
\caption{Results on the MIMIC-3 (\%). $^\dag$ indicates the improvement achieved by the proposed rescoring framework. The best scores for each base predictor are marked in bold.}
\label{tab:mimic3} 
\end{table*}

\begin{table}[t!]
\begin{center}
\begin{tabular}{lcc}
\toprule
 \bf Model & \bf Macro F1 & \bf Micro F1 \\
\midrule
CAML & 4.90 & 44.79 \\
 ~+ MADE  & 5.31$^\dag$ & \bf 46.11$^\dag$ \\
 ~+ Mask-SA & \bf 5.55$^\dag$ & 46.08$^\dag$ \\
\midrule
MultiResCNN & 5.06 & 45.89 \\
 ~+ MADE & 5.50$^\dag$ & 47.49$^\dag$ \\
 ~+ Mask-SA & \bf 5.88$^\dag$ & \bf 47.55$^\dag$ \\
\midrule
LAAT & 6.41 & 47.54 \\ 
~+ MADE  & 7.23$^\dag$ & \bf 49.15$^\dag$ \\ 
~+ Mask-SA & \bf 7.42$^\dag$ & 49.05$^\dag$ \\
\bottomrule
\end{tabular}
\end{center}
\caption{Results on the MIMIC-2 test set using all codes (\%). $^\dag$ indicates the improvement achieved by the proposed framework. The best scores are marked in bold.}
\label{tab:mimic2} 
\vspace{-2mm}
\end{table}

\subsection{Setup}
We evaluate our model on two benchmark datasets for ICD code prediction.
\begin{compactitem}
    \item \textbf{MIMIC-2} \quad Following the prior work~\cite{mullenbach-etal-2018-explainable,li2020multirescnn}, we evaluate our method on the MIMIC-2 dataset.
    We follow their setting, where 20,533 summaries are used for training, and 2,282 summaries are used for testing.
    There are 5,031 labels in the dataset.
    \item \textbf{MIMIC-3} \quad The Medical Information Mart for Intensive Care III (MIMIC-3)~\cite{johnson2016mimic} dataset is a benchmark dataset which contains text and structured records from a hospital ICU.
    We use the same setting as the prior work~\cite{mullenbach-etal-2018-explainable}, where there are 47,724 discharge summaries for training, with 1,632 summaries and 3,372 summaries for validation and testing, respectively.
    There are 8,922 labels in the dataset.
    We also follow the setting in~\cite{shi2017towards} where only the top-50 most frequent codes are considered. This setting has 8,067 summaries for training, 1,574 summaries for validation, and 1,730 summaries for testing.
\end{compactitem}

We follow the preprocessing steps described in~\citet{mullenbach-etal-2018-explainable} with the provided scripts~\footnote{\url{https://github.com/jamesmullenbach/caml-mimic}}.
All discharge summaries are truncated to a maximum length of 2,500 tokens.

\subsection{Base Predictors}
In order to validate the generalizability of our proposed framework, we employ three different base predictors that are proposed in prior work:
\begin{compactitem}
    \item \textbf{CAML} \quad Convolutional attention for multi-label classification (CAML) is a method proposed in~\cite{mullenbach-etal-2018-explainable}. CAML aimed at improving ICD code prediction by exploiting the textual description of codes with attention mechanism~\cite{bahdanau2014neural}.
    \item \textbf{MultiResCNN} \quad The multi-filter residual convolutional neural network (MultiResCNN) improved the design of CAML with multiple convolution filters and residual connections~\cite{li2020multirescnn}.
    \item \textbf{LAAT} \quad ~\citet{ijcai2020-461} proposed a label attention model which augments the label attention mechanism in CAML with additional transformations. It achieved state-of-the-art performance on MIMIC-2 and MIMIC-3 datasets.
\end{compactitem}


\subsection{Training and Evaluation Details}
We train our rerankers with the label sets in the training set for 30 epochs. 
Adam is chosen as the optimizer with a learning rate of $2e-5$. The batch-size is set to 64. 
The MADE reranker has one hidden layer with 500 neurons, and we find that using $n = 10$ different orderings provides good estimation without using too much computation power.
The Mask-SA reranker employs the transformer architecture with 6 self-attention layers, each with 8 attention heads. The hidden size is set to 256.
For each pair of the base predictor and the reranker, we apply a grid search over possible values of $\alpha$ and $\beta$ on the validation set to find the best-performing hyperparameters, and we use them to perform evaluation on the test set.
During reranking, we generate top-50 label set candidates to rerank.
Note that our approach is to rerank the label set candidates instead of modifying the predicted probabilities from the base predictors.
Therefore, common metrics considering the predicted probabilities of each label, such as Precision@K and AUC, are not suitable for our evaluation.
Instead, we evaluate our methods with two metrics, macro F1 and micro F1.

\subsection{Results}
The results on the MIMIC-3 and MIMIC-2 datasets are shown in Table~\ref{tab:mimic3} and Table~\ref{tab:mimic2} respecitively.
All results are obtained by averaging the scores of 5 different training runs.
We list the results before reranking in the first row of each base predictor.

In all scenarios, our proposed reranking framework achieves consistent improvement over the base predictors except for LAAT, where the macro F-score on MIMIC-3 top-50 dev set slightly decreased.
The relative improvement in macro F-score for all-code settings are more significant, ranging from 1\% to 16\%.
Considering that the all-code setting is much more challenging and macro F-score is difficult to improve due to the data imbalance issue, the achieved improvement demonstrates the great potential of the proposed framework for better practicality.
The MADE and Mask-SA reranker are both effective for the purpose of reranking.
Their gains are similar across different settings and datasets.
We believe that this trend is reasonable given that their formulations are similar, i.e., they both calculate score as product of conditional probabilities.
We also observe that in the settings using all ICD codes, Mask-SA reranker provides larger improvement to the macro F-score consistently.

\begin{figure}[t!]
\centering
\includegraphics[width=\linewidth]{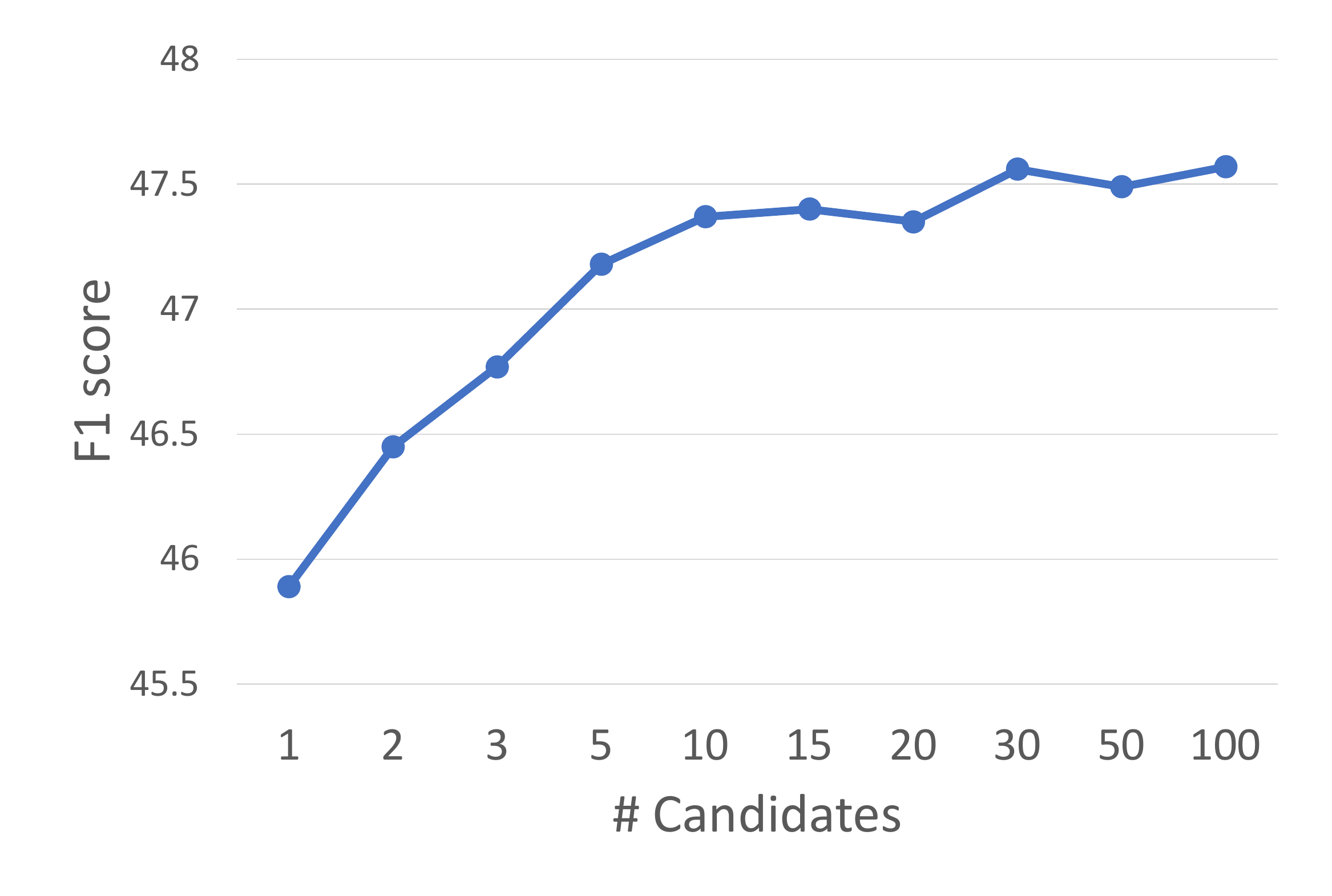}
\vspace{-5mm}
\caption{F1 scores (\%) with different number of candidates.}
\label{fig:nbest}
\vspace{-2mm}
\end{figure}

Our proposed framework improves upon the best-performing methods on all settings.
Note that the proposed framework is complementary to the base predictor.
The results show that our reranking method can improve upon any predictor that is designed with the independent assumption, demonstrating the great flexibility and generalizability of our method.

\subsection{Effect of Candidate Numbers}
The reranking results reported in Table~\ref{tab:mimic3} and Table~\ref{tab:mimic2} are generated with top-50 candidates.
In order to investigate the effect of number of candidates to the final performance, we plot the performance with regard to different number of candidates in Figure~\ref{fig:nbest}.
As shown in the figure, the reranked score increases consistently when the number of candidates is less than 10. No significant improvement is observed when the number of candidates is larger than 10.

We hypothesize that this phenomenon is due to our formulation of the final score.
When calculating the final scores, we combine the original score from the base predictor and the score from the reranker.
For the candidates that originally ranked after 10 by the base predictor, the original score may be too low; hence it is almost impossible to be selected after reranking.

\subsection{Effectiveness of Reranking}
The ultimate goal of our reranker is to bring the best-performing label set to the highest rank.
In order to further examine the effectiveness of our rerankers, we calculate the average ranking of the best-performing label set, i.e. the set with the highest micro F-score with respect to the ground truth, before and after reranking.
The results are shown in Table~\ref{tab:oracle-ranking}, implying that the proposed model can bring the best candidate from the 24-th place to the 19-th place for better practicality in terms of the systems with doctors' interactions.
Our rerankers improve the average ranking by more than 20\% relative, demonstrating that the reranking process is effective.

\begin{table}[t!]
\begin{center}
\begin{tabular}{lcc}
\toprule
 \bf Model & \bf Avg. Rank  \\
\midrule
LAAT & 24.58  \\
 ~+ MADE  & 19.73 \\
 ~+ Mask-SA & 19.50 \\
\bottomrule
\end{tabular}
\end{center}
\caption{Average rank of the best-performing label set among the top-50 candidates.}
\label{tab:oracle-ranking} \vspace{-2mm}
\end{table}

\begin{figure}[t!]
\centering
\vspace{2mm}
\includegraphics[width=\linewidth]{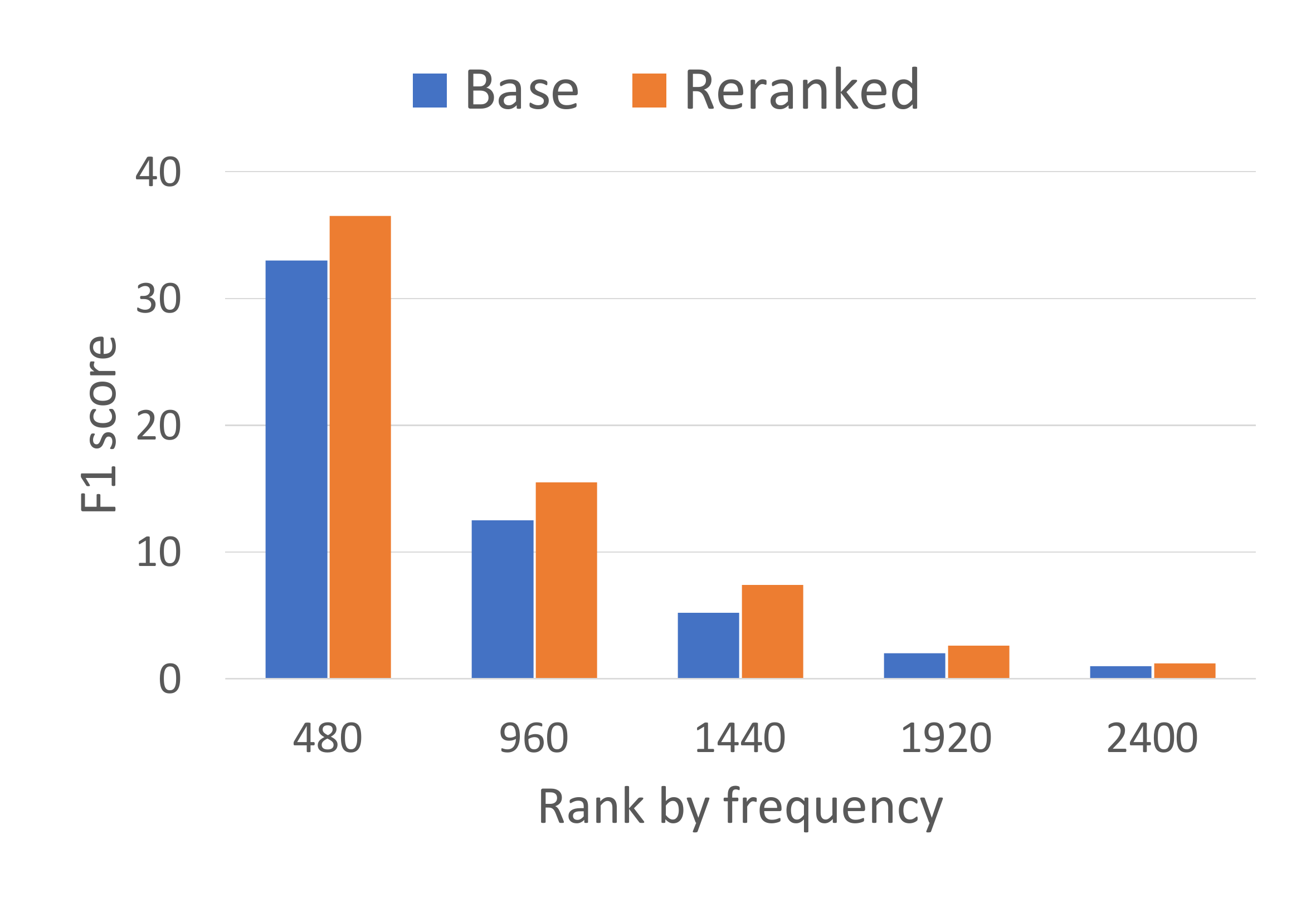}
\caption{F1 scores (\%) with regard to the frequencies of labels. We bucket labels by their frequencies into 6 buckets.}
\label{fig:infrequent}
\end{figure}

\begin{table*}[t!]
\begin{center}
\begin{tabular}{ll}
\toprule
& \bf Baseline + Rescoring\\
\midrule
Sample 1  & 427.1 427.41 427.5 693.0 99.6 995.0\\
 (+ Mask-SA) & 427.1 427.41 427.5 693.0 99.6 995.0 99.62 {\color{blue}96.04 96.71}    \\
\midrule
Sample 2 & 571.5 733.00 733.09 96.04 96.72 V66.7 \\
(+ MADE) & 571.5 733.00 {\color{red}\sout{733.09}} 96.04 96.72 V66.7 305.1 431 96.6 \\
\bottomrule
\end{tabular}
\end{center}
\vspace{-2mm}
\caption{Sample results before and after reranking from MIMIC-3 data.}
\label{tab:sample} 
\vspace{-2mm}
\end{table*}

\subsection{Effect on Infrequent Labels}
The task of ICD code prediction is extremely hard due to the large set of labels and the imbalanced data: the top-50 most frequent codes take up more than a third of all the outputs.
To investigate the effect of the proposed framework on the infrequent labels, we bucket the labels according to their frequencies, and calculate the performance for each bucket.
We plot the performance of MultiResCNN on MIMIC-3 full set with regard to label frequencies in Figure~\ref{fig:infrequent}.
The figure demonstrates that with reranking, the performance of the infrequent labels also increases.
This result indicates that the reranking method is helpful for the extreme multi-label classification problem.

\section{Qualitative Analysis}
After comparing the results produced from two rerankers, we find that both methods have similar tendency of prediction. In other words, the original candidate sets would be improved by adding or deleting similar ICD codes after rescoring from both MADE and Mask-SA.
To further analyze prediction change, Table~\ref{tab:sample} shows the original and reranked results for two data samples.

\subsection{Addition and Deletion of Predictions}
For the first sample, we find that the reranking module tends to add missing ICD codes to the predicted set.
Specifically, the first sample has no 96 category in the original prediction, and the rescoring process adds 96.04 and 96.71 (highlighted in blue) in the candidate set for better performance.
By checking their meanings, we could know that 96.04 is about insertion of endotracheal tube and 96.71 is about invasive mechanical ventilation, and both treatments are important for patients in ICU maintaining their respiratory function. Due to their strong dependency, we find that these codes frequently co-occur in the training data. Apparently, the reranker learn the correlation and is capable of improving the prediction in terms of both diversity and accuracy.

In the second sample, it can be found that our module can also help remove the unreasonable codes.
Specifically, the code 733.09 (highlighted in red) is not proper to be the selected code due to the appearance of 733.00, which is the correct disease from the record.
Therefore, the reranker can help not only provide additional accurate codes but also delete unreasonable ones for better performance.

\subsection{Reranking Analysis}
We further analyze our methods from the top-10 ranking candidates sets in the second sample to confirm if the sets with more accurate ICD codes would be at the top of the reranked sets.
In this sample, the unreasonable ICD code 733.09 appears in every top-10 predictions before reranking.
With reranking, our reranker is able to bring the set without 733.09 to the top-1.
This example demonstrates that the reranker's ability to identify conflicting predictions and that we are able to correct them with the proposed framework.



\section{Conclusions}
This paper proposes a novel framework to improve multi-label classification for automatic ICD coding, which includes candidate generation and candidate reranking modules. In the first stage, a base predictor is performed to generate top-k probable label set candidates. In the second stage, we propose a reranker to capture the correlation between ICD labels without any external knowledge. Two types of the reranker, MADE and Mask-SA, are employed to rerank the candidate sets. 
Our experiments show that both rerankers can consistently improve the performance of all predictors in MIMIC-2 and MIMIC-3 datasets, demonstrating the generalizability of our framework and the great potential of the flexible usage.

\bibliography{naacl2021}
\bibliographystyle{acl_natbib}




\end{document}